\documentclass[sigconf,nonacm]{acmart}
\usepackage[ruled]{algorithm2e}
\usepackage{caption}
\usepackage{subcaption}
\usepackage{multirow}

\AtBeginDocument{%
  \providecommand\BibTeX{{%
    \normalfont B\kern-0.5em{\scshape i\kern-0.25em b}\kern-0.8em\TeX}}}

\setcopyright{acmcopyright}
\copyrightyear{2023}
\acmYear{2023}
\acmDOI{XXXXXXX.XXXXXXX}

\acmConference[AIMLSystems 2023]{ The Third International Conference on Artificial Intelligence and Machine Learning Systems}{October 25--28, 2023}{Bangalore, India}

\acmBooktitle{The Third International Conference on AI-ML Systems (AIMLSystems 2023), October 25--28, 2023, Bangalore, India}
\acmPrice{15.00}
\acmISBN{978-1-4503-XXXX-X/18/06}

\renewcommand\footnotetextcopyrightpermission[1]{} 

\begin{document}

\title[TIFeD: a Tiny Integer-based Federated learning algorithm with Direct feedback alignment]{TIFeD: a Tiny Integer-based Federated learning algorithm \\ with Direct feedback alignment}

%


\author{Luca Colombo}
\affiliation{%
  \institution{Politecnico di Milano}
  \streetaddress{Milan, Italy}
  \city{Milan}
  \country{Italy}}
\email{luca2.colombo@polimi.it}

\author{Alessandro Falcetta}
\affiliation{%
  \institution{Politecnico di Milano}
  \streetaddress{Milan, Italy}
  \city{Milan}
  \country{Italy}}
\email{alessandro.falcetta@polimi.it}

\author{Manuel Roveri}
\affiliation{%
  \institution{Politecnico di Milano}
  \streetaddress{Milan, Italy}
  \city{Milan}
  \country{Italy}}
\email{manuel.roveri@polimi.it}

\thanks{© ACM 2023. This is the author's version of the work. It is posted here for your personal use. Not for redistribution. The definitive Version of Record was published in the \textit{Proceedings of the Third International Conference on AI-ML Systems (AIMLSystems 2023)}, \href{http://dx.doi.org/10.1145/3639856.3639867}{http://dx.doi.org/10.1145/3639856.3639867}.}


\begin{abstract}
Training machine and deep learning models directly on extremely resource-constrained devices is the next challenge in the field of tiny machine learning. The related literature in this field is very limited, since most of the solutions focus only on \textit{on-device} inference or model adaptation through online learning, leaving the training to be carried out on external Cloud services. An interesting technological perspective is to exploit Federated Learning (FL), which allows multiple devices to collaboratively train a shared model in a distributed way. However, the main drawback of state-of-the-art FL algorithms is that they are not suitable for running on tiny devices. For the first time in the literature, in this paper we introduce \alg{}, a Tiny Integer-based Federated learning algorithm with Direct Feedback Alignment (DFA) entirely implemented by using an integer-only arithmetic and being specifically designed to operate on devices with limited resources in terms of memory, computation and energy. Besides the traditional \textit{full-network} operating modality, in which each device of the FL setting trains the entire neural network on its own local data, we propose an innovative \textit{single-layer} \alg{} implementation, which enables each device to train only a portion of the neural network model and opens the door to a new way of distributing the learning procedure across multiple devices. The experimental results show the feasibility and effectiveness of the proposed solution. The proposed \alg{} algorithm, with its \textit{full-network} and \textit{single-layer} implementations, is made available to the scientific community as a public repository.
\end{abstract}




\def\alg{$\mathsf{TIFeD}$}
\def\boldalg{$\boldsymbol{\mathsf{TIFeD}}$}

\maketitle

\section{Introduction}
\label{sec:introduction}
Tiny Machine Learning (TinyML) is a new emerging research area aiming at bringing Machine Learning (ML) and Deep Learning (DL) algorithms on embedded systems and Internet-of-Things (IoT) devices~\cite{banbury2020benchmarking}. The main reason behind this paradigm is that, in recent years, the scientific trend is to move the processing of data as close as possible to where they are generated. This leads to several advantages. First, it enables decision making directly on devices, hence reducing the latency between data production and data processing and supporting real-time applications. Second, it increases the energy efficiency, given that sending data is more power-consuming than performing computations on device. Third, it enhances privacy and security because sensitive data remain on the IoT units and are not transmitted to remote servers~\cite{stankovic1996real}.

The main drawback of TinyML, however, is that on-device training of DL models is, in most cases, challenging or even impossible due to the severe technological constraints on memory, computation and energy characterizing IoT units. In the related literature, only a few works, such as~\cite{sundaramoorthy2018harnet, disabato2020incremental}, have attempted to implement an incremental mechanism, based on transfer learning, to locally adapt the model as new data become available.




To provide on-device training within an IoT system, an interesting technological perspective is to leverage Federated Learning (FL), which is a distributed approach that allows multiple devices to collaboratively train a shared model. Each device performs a local training with its own data, and then sends the model updates to a central server that aggregates the locally computed updates of all the devices in the network~\cite{bonawitz2019towards}. The main drawback of state-of-the-art FL algorithms is that they are designed to operate on edge devices such as smartphones and tablets, which are significantly more powerful from the technological point of view than IoT units. For this reason, traditional FL is not suitable for running on resource-constrained devices, given that: \textit{(i)} the traditional learning procedure with Backpropagation (BP) may not be feasible since it is computationally expensive; \textit{(ii)} the Neural Network (NN) must be shallow and with few parameters; \textit{(iii)} the training dataset has to be small enough to fit into the device's memory.

In this perspective, the aim of this paper is to address the following research question: \textit{how can we train deep and complex machine learning models directly on extremely resource-constrained devices in a federated way? } To the best of our knowledge, we propose for the first time in the literature a Tiny Integer-based Federated learning algorithm with Direct feedback alignment (\alg) that introduces the following innovations:
\begin{itemize}
\item It enables resource-constrained devices to train a NN without having to rely on external Cloud services by leveraging federated learning. In this way, we can exploit all the advantages of TinyML also during the learning phase and not only for the inference.

\item It allows only single layers to be trained, thus reducing the computational and memory demands on the micro-controllers. A direct consequence is that sending weights updates also requires less data since only a portion of the model needs to be sent to the central server, resulting in energy savings.

\item It is entirely implemented through an integer-only arithmetic. In this way, memory consumption can be reduced by using integers less than 32 bits long and, in addition, even devices without floating-point unit (FPU) can train ML and DL models.
\end{itemize}
The proposed \alg{} algorithm is based on the Direct Feedback Alignment (DFA)~\cite{nokland2016direct} learning procedure, which has two main advantages over BP. First, it is much less computationally expensive because each hidden layer is trained independently from the rest of the NN. Second, it allows the implementation of an integer-only arithmetic while avoiding the risk of overflow, which is a major issue when using BP with integer operations.

The paper is organized as follows. Section~\ref{sec:related_literature} gives an overview on the related literature. In Section~\ref{sec:background}, the background on DFA and FL is introduced. Section~\ref{sec:solution} presents the proposed \alg{} algorithm, while experimental results are shown and discussed in Section~\ref{sec:results}. Conclusions are finally drawn in Section~\ref{sec:conclusions}.

\section{Related Literature}
\label{sec:related_literature}
This section describes the related literature in the field of federated learning and the available works aiming at implementing the most used FL algorithm, namely Federated Averaging (FedAvg), on tiny devices.

The most popular FL algorithm, introduced by Google in 2016, is Federated Averaging~\cite{mcmahan2017communication}. In FedAvg, each node performs multiple iterations of mini-batch Stochastic Gradient Descent (SGD) with BP to update the local model before sending the gradients to the server. In the central server, received model updates are aggregated together by using an averaging function to combine knowledge learned from different datasets. The main disadvantage of FedAvg is that it has not been designed to operate on resource-constrained devices, as it requires a large amount of memory and high processing power to perform the local learning procedure, and thus it is not suitable in a TinyML scenario. Other works in the field, such as~\cite{yang2020federated, konevcny2016federated, goetz2019active}, try to optimize the FedAvg algorithm either by improving its efficiency or by using more complex and sophisticated aggregation functions, but without addressing the issue of TinyML-specific design.

On the other hand, \cite{llisterri2022device, kopparapu2021tinyfedtl} are the first papers implementing FedAvg on a device with limited resources, the Arduino Nano 33 BLE Sense board. In particular, Kopparapu et al.~\cite{kopparapu2021tinyfedtl} exploited transfer learning to solve a binary image classification task, while Llisterri et al.~\cite{llisterri2022device} implemented a feed-forward neural network on a simpler keyword spotting task. Given the aforementioned limitations of FedAvg in working on tiny devices, however, both works are only able to train the last fully-connected layer of the considered neural networks.

Differently from the literature, the solution proposed in this paper provides some fundamental aspects for on-device training of ML models, as summarized in Table~\ref{tab:comparison}. In particular, \alg{} is specifically designed for operating on resource-constrained devices and it is entirely implemented using an integer-only arithmetic. Moreover, the \textit{single-layer} \alg{} implementation introduces a new operating modality in which each device can train only a portion of the global shared model.

\begin{table}[t]
\caption{Comparison with existing solutions.}
\label{tab:comparison}
\begin{tabular}{@{}cccc@{}}
\toprule
    Algorithms & \begin{tabular}{@{}c@{}}Tiny devices- \\ specific design \end{tabular} & \begin{tabular}{@{}c@{}}Integer-only \\ arithmetic \end{tabular} & \begin{tabular}{@{}c@{}}Partial model \\ training \end{tabular} \\
    \midrule
    \cite{goetz2019active} & No & No & No \\
    \cite{kopparapu2021tinyfedtl} & Yes & No & No \\
    \cite{mcmahan2017communication} & No & No & No \\
    \boldalg{} & \textbf{Yes} & \textbf{Yes} & \textbf{Yes} \\
    \bottomrule
\end{tabular}
\end{table}

\section{Background}
\label{sec:background}
This section introduces the basic concepts on both the DFA training algorithm and the Federated Learning approach.

\subsection{Direct Feedback Alignment (DFA)}
\label{subsec:dfa}
Direct Feedback Alignment~\cite{nokland2016direct} is a new learning method for ML and DL models based on the Feedback Alignment (FA)~\cite{lillicrap2014random} principle. FA showed that the symmetry of weights used in the forward and backward phases, which is required in the BP algorithm, is not necessary. In simpler terms, the network can learn to effectively use fixed and random feedback weights to reduce the error.

The FA training algorithm is based on two insights~\cite{lillicrap2014random}: 
\begin{enumerate}
\item The feedback weights do not need to be exactly equal to the forward weights $W$, but it is sufficient that, for any random matrix $B$ whose elements are uniformly distributed, on average
\begin{flalign}\label{eq:fa}
&\textbf{e}^{\intercal}W B\textbf{e} > 0
\end{flalign}
where $\textbf{e}$ is the error of the network.

\item In order to guarantee Eq.~(\ref{eq:fa}), instead of adjusting $B$, we can keep it fixed and adjust $W$ accordingly. From a geometrical point of view, Eq.~(\ref{eq:fa}) means that the teaching signal used by FA (i.e., $B\textbf{e}$) lies within 90° of the signal used by BP (i.e., $\textbf{e}^{\intercal} W$), that is, $B$ pushes the weights in roughly the same direction as BP. This alignment of signals implies that $B$ acts as $W^\intercal$ and, since $B$ is fixed, the alignment is driven by changes in the forward weights $W$. In this way, even random feedback weights convey useful teaching information throughout the network.
\end{enumerate}

When adopted in deep networks with more than one hidden layer, however, even FA back-propagates the error from the output layer through the upper layers. On the other hand, DFA propagates the output error directly to each of the hidden layers and exploits the FA principle to train them independently of the rest of the neural network.

Let $H$ be the number of hidden layers of a feed-forward NN. Formally, the layer's update directions for FA -- denoted by $\delta_{FA}^h$, $h=1, \dots, H$ -- are calculated as:
\begin{flalign*}
&\delta_{FA}^h \quad = \quad \left\{ \begin{array}{rcl} 
L^\prime\odot \mathsf{act}^{\prime\,h}(\textbf{a}^h) & \mbox{for} & h=H \\
\delta_{FA}^{h+1}B^{h}\odot \mathsf{act}^{\prime\,h}(\textbf{a}^h) & \mbox{for} & 1\leq h\leq H-1
\end{array}\right.
\end{flalign*}
where $L^\prime$ is the derivative of the Loss Function, and $\mathsf{act}^{\prime\,h}(\cdot)$, $\textbf{a}^{h}$, and $B^{h}$ are the derivative of the activation function, the sum of products, and the fixed random weight matrix of layer $h$, respectively.

Concerning DFA, the layer's update directions -- denoted by $\delta_{DFA}^h$, $h=1, \dots, H$ -- are computed as:
\begin{flalign}\label{eq:dfa}
\delta_{DFA}^h = L^\prime B^h\odot \mathsf{act}^{\prime\:h}(\textbf{a}^h).
\end{flalign}

Once $\delta_{DFA}^h$, with $h=1, \dots, H$, have been computed, the weights updates for each layer are computed as:
\begin{flalign}
\label{eq:weights_update}
\begin{split}
&\delta {W}^h = - \textbf{o}^{h-1\:\intercal}\delta_{DFA}^h \\
&\delta \textbf{b}^h = - \delta_{DFA}^h
\end{split}
\end{flalign}
where $\textbf{o}^{h-1}$ is the output of the previous layer in the network.


According to~\cite{refinetti2021align}, a network trained with DFA converges to the same region of the loss landscape regardless of the initialization of the network weights $W$. This property is guaranteed by the alignment step of the DFA learning procedure, in which the weights $W$ align with the random feedback vectors $B$ used to propagate the network error to each layer. The alignment phase also explains why the random feedback vectors $B$ must be considered as an initialization of the network. Indeed, they drive the learning towards different minima of the loss function, resulting in different performance and accuracies of the final model. DFA has been successfully used in PocketNN~\cite{song2022pocketnn}, a feed-forward neural network tailored for IoT devices.

\subsection{Federated Learning}
\label{subsec:fl}
Federated learning allows multiple clients to train a shared ML model on decentralized data sources. It addresses the drawbacks of classical ML (i.e. data privacy and security) by distributing the learning procedure across several nodes, thus allowing the data to remain local. Instead of collecting and sending acquired data to an external Cloud service, each device performs a local training by using a shared model as initialization and exchanges only the weights updates with a central server. Local updates are then aggregated by the server to create a global model that will be forwarded to the clients. This process is repeated iteratively until the global model reaches the desired level of accuracy~\cite{konevcny2016federated, kairouz2021advances}.

\section{The Proposed Solution}
\label{sec:solution}
This section details the proposed \alg{} algorithm. We emphasize that, as stated in Section~\ref{sec:related_literature} and outlined in Table~\ref{tab:comparison}, our work introduces the following innovations compared with the literature. First, \alg{} is specifically designed to operate on resource-constrained devices, taking into account their technological limitations in terms of memory, computation and energy. Second, inspired by~\cite{song2022pocketnn}, \alg{} is implemented using an integer-only arithmetic, which enables a reduction in memory consumption by using integers less than 32 bit long and, in addition, allows DL models to be trained on devices without a floating-point unit. Third, the \textit{single-layer} \alg{} implementation introduces a new way of distributing the learning procedure across multiple devices, allowing each IoT unit to train only a portion of the whole NN.

In particular, Subsection~\ref{subsec:overview} gives a general overview on the proposed federated learning scenario. In Subsection~\ref{subsec:feddfa}, the transition from DFA to Federated DFA is described, while Subsection~\ref{subsec:single_layers} presents an extension to Federated DFA in which each client node trains only specific layers of the considered neural networks.

\subsection{Overview}
\label{subsec:overview}

\begin{figure}[t]
  \centering
  \includegraphics[width=\linewidth]{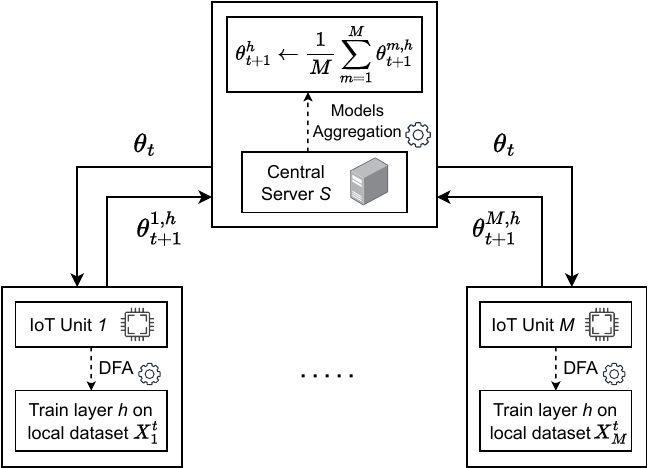}
  \caption{Overview of the \alg{} algorithm operating in a federated learning scenario.}
  \label{fig:architecture}
\end{figure}

An overview of the proposed \alg{} algorithm operating in a FL scenario is shown in Fig.~\ref{fig:architecture}. Without any loss of generality, in the followings a supervised learning image classification task is considered as a reference.

Consider a pervasive system consisting of $M$ IoT units, each of which collects local data $X_m$ through some sensors. The objective is to obtain a ML model trained directly on-device by the IoT units, without relying on external Cloud services. To achieve this goal, the proposed \alg{} algorithm, detailed in the next subsections, exploits FL. The key role in this setup is played by the central server $S$, which is the orchestrator of the entire architecture. At each communication round $t$, the server $S$ sends the global model $\theta_t$ to all the $M$ IoT units, that perform a local computation of the DFA algorithm based on their dataset $X_m^t$. Once the learning procedure is completed, each IoT unit $m$, with $m=1, \dots, M$, submits the local updates $\theta_{t+1}^m$ to the server $S$, which is in charge of aggregating the results into a new global model.

In particular, the proposed \alg{} algorithm can operate in two modalities. The first, called \textit{full-network} implementation (detailed in Subsection~\ref{subsec:feddfa}), allows each IoT unit to train all the $H$ hidden layers of the DL model. In this case, the result of the computation will be a set containing the weights and bias updates for each layer $h=1,\dots,H$. The second, named \textit{single-layer} implementation (explained in Subsection~\ref{subsec:single_layers}), enables the IoT units to train only the $h$-th layer of the NN. The result of the computation, in this second scenario, will be a single tuple containing the weights and bias updates related to layer $h$.

\begin{algorithm}[t]
\DontPrintSemicolon
\SetKwProg{FSer}{central server executes}{:}{}
\SetKwProg{FNode}{NodeUpdate}{$(m, t, \theta)$:}{}
\SetKwFor{For}{foreach}{do}{end for}
\SetKwFor{ForPrl}{foreach}{in parallel do}{end for}
    \FSer{}{
        initialize $B^h$ \textbf{foreach} $h = 1, \dots, H$ \;
        \For{round $t = 1,\dots,T$}{
            \ForPrl{device $m = 1,\dots,M$}{
                $\theta_{t+1}^m \longleftarrow$ NodeUpdate($m, t, \theta_t$) \;
            }
            $\theta_{t+1} \longleftarrow \frac{1}{M} \sum_{m=1}^{M} \theta_{t+1}^m$ \;
        }
    }
    \;
    \FNode{}{
        $\mathcal{A} \longleftarrow$ (split $X_m^t$ into mini-batches of size $bs$) \;
        \For{local epoch $e$}{
            \For{mini-batch $\alpha \in \mathcal{A}$}{
                $\delta \theta \longleftarrow \left\{[\delta {W}^1, \delta \textbf{b}^1], \: \dots \: , [\delta {W}^H, \delta \textbf{b}^H]\right\}$ \;
                $\theta \longleftarrow \theta - \eta \: \delta \theta$ \;
            }
        }
        return $\theta_{t+1}^m$ to the central server \;
    }
\caption{   
\textit{Full-network} \alg{} algorithm implementation. $M$ is the number of devices, $H$ is the number of hidden layers, $bs$ is the mini-batch size, $X_m^t$ is the local dataset at the $t$-th round of client $m$, $e$ is the number of local epochs, and $\eta$ is the learning rate.
\label{alg:tifed}
}
\end{algorithm}

\begin{algorithm}[t]
\DontPrintSemicolon
\SetKwProg{FSer}{central server executes}{:}{}
\SetKwProg{FNode}{NodeUpdate}{$(m, t, h, \theta)$:}{}
\SetKwFor{For}{foreach}{do}{end for}
\SetKwFor{ForPrl}{foreach}{in parallel do}{end for}
    \FSer{}{
        initialize $B^h$ \textbf{foreach} $h = 1, \dots, H$ \;
        $s \longleftarrow \left \lfloor{\frac{M}{H}}\right \rfloor$ \;
        \For{round $t = 1,\dots,T$}{
            $\mathcal{S}_h \longleftarrow$ (split the clients into $H$ groups of size $s$) \;
            \ForPrl{group $h = 1,\dots,H$}{           
                \ForPrl{device $m \in \mathcal{S}_h$}{
                    $\theta_{t+1}^{m,h} \longleftarrow$ NodeUpdate($m, t, h, \theta_t$) \;
                }
                $\theta_{t+1}^{h} \longleftarrow \frac{1}{s} \sum_{m=1}^{s} \theta_{t+1}^{m,h}$ \;
            }
            $\theta_{t+1} \longleftarrow \left\{\theta_{t+1}^{1}, \: \dots \: , \theta_{t+1}^{H}\right\}$ \;
        }
    }
    \;
    \FNode{}{
        $\mathcal{A} \longleftarrow$ (split $X_m^t$ into mini-batches of size $bs$) \;
        \For{local epoch $e$}{
            \For{mini-batch $\alpha \in \mathcal{A}$}{
                $\delta \theta^h \longleftarrow \left\{[\delta {W}^h, \delta \textbf{b}^h]\right\}$ \;
                $\theta^h \longleftarrow \theta^h - \eta \: \delta \theta^h$ \;
            }
        }
        return $\theta^{m,h}_{t+1}$ to the central server \;
    }
\caption{   
\textit{Single-layer} \alg{} algorithm implementation. $M$ is the number of devices, $H$ is the number of hidden layers, $bs$ is the mini-batch size, $X_m^t$ is the local dataset at the $t$-th round of client $m$, $e$ is the number of local epochs, and $\eta$ is the learning rate.
\label{alg:sltifed}
}
\end{algorithm}

\subsection{From DFA to Federated DFA}
\label{subsec:feddfa}
In the federated learning setting, we consider a fixed set of clients $M$, each with a local dataset $X_m$, and a global model $\theta$ composed of $H$ hidden layers. At the beginning of each round $t$, the central server sends the current model parameter 
\begin{flalign*}
\theta_t = \left\{[W_t^1, \textbf{b}_t^1], \: \dots \: , [W_t^H, \textbf{b}_t^H]\right\},
\end{flalign*}
being $W_t^h$ and $\textbf{b}_t^h$ the matrix of weights and the bias vector of layer $h$ respectively, to each of the clients, which perform a local computation of the DFA algorithm. Once the weight updates have been computed by each client $m$ on its dataset $X_m^t$, the new local model parameters $\theta_{t+1}^m$ are sent to the central server, which aggregates all the results and updates its global state with the new model $\theta_{t+1}$.

It is important to note that since we are in an embedded scenario where devices have a limited amount of memory (e.g., the Arduino Nano 33 BLE Sense board has 256kB of SRAM), the local dataset $X_m$ cannot be arbitrarily big, because it must fit into the device's memory. This is a major limitation compared to traditional FL, where learning procedures benefit from very large and varied datasets.

More in detail, the \textit{full-network} \alg{} algorithm implementation, summarized in Algorithm~\ref{alg:tifed}, can be divided into two blocks: the central server component and the IoT units component. Concerning the former part, executed by the \textbf{central server}, it works as follows:
\begin{enumerate}
\item The node initializes, for each hidden layer $h$ of the network, the matrix of feedback weights $B^h$ by sampling values from a uniform distribution, and sends them to each client $m$.
\item For each round $t = 1, \dots, T$, the node sends the current model parameters $\theta_t$ to the clients in order to let them perform local updates using DFA.
\item After receiving the $M$ local updates $\theta_{t+1}^m$ from the clients, the node upgrades the global model $\theta_{t+1}$ by aggregating the results with the average function:
\begin{align*}
\theta_{t+1} \longleftarrow \frac{1}{M} \sum_{m=1}^{M} \theta_{t+1}^m.
\end{align*}
\end{enumerate}

Regarding the \textbf{NodeUpdate} function, executed by each IoT unit, it works as follows:
\begin{enumerate}
\item The node splits its local part of the dataset $X_m^t$ (relative to the current round $t$) into mini-batches $\alpha$ of size $bs$.
\item For each local epoch $e$ and for each mini-batch $\alpha$, the node computes the weights updates through the DFA learning algorithm, as shown in Eq.~(\ref{eq:weights_update}), and applies them to the current model parameters $\theta_{t}$.
\item Once the computation is completed, the node sends the new model parameters $\theta_{t+1}$ back to the central server.
\end{enumerate}

The models aggregation can be done at different points: after each mini-batch $\alpha$, after the entire buffer $X_m^t$, or after the number of training passes $e$ each client makes on the current buffer $X_m^t$. The choice of the aggregation point represents a trade-off between the accuracy of the final model and the number of required communication rounds, which results into higher power consumption. A detailed analysis about this point is provided in the experimental results in Section~\ref{sec:results}.

\subsection{Single layers training}
\label{subsec:single_layers}
One of the main advantages of the DFA learning algorithm is the ability to train each NN layer independently. Indeed, as shown in Eq.~(\ref{eq:dfa}), the layer's update direction formula $\delta_{DFA}^h$ is not recursive and, for each hidden layer $h$, depends only on the derivative of the Loss Function $L^\prime$, the fixed matrix of feedback weights $B^h$ and the derivative of the activation function $\mathsf{act}^{\prime\,h}(\cdot)$. This means that, unlike BP, the network error is propagated directly to each hidden layer and does not have to traverse the entire network. 

By exploiting this feature, it is possible to design a federated algorithm in which each IoT unit, instead of training an entire network on its local dataset, trains only a single layer of the global model. This results in two benefits: \textit{(i)} the algorithm is more lightweight in terms of computational and memory demands and can therefore run on devices with tighter resource constraints, and \textit{(ii)} the communication between the devices and the central server requires the exchange of a smaller amount of data since only a portion of the network (i.e., only the updates $\theta_{t+1}^{m,h}$ of the $h$-th layer) needs to be sent, resulting in energy savings. On the other hand, the drawbacks are that some overhead is added to the central server that has to decide which device trains which layer, and that there is a loss in the accuracy of the final global model. More accurate analyses are provided in Section~\ref{sec:results}.

The \textit{single-layer} \alg{} algorithm implementation is shown in Algorithm~\ref{alg:sltifed} and works as follows:
\begin{itemize}
\item The central server also defines, for each round $t$, the set $\mathcal{S}_h$ which contains a random splitting of the clients into $H$ different groups of size $s \longleftarrow \left \lfloor{\frac{M}{H}}\right \rfloor$. Each group is in charge of training only the $h$-th layer of the network.
\item The IoT units compute the weights update related only to the $h$-th layer through the DFA learning algorithm, apply it to the current model parameters $\theta_t^h$, and send the new model parameters $\theta_{t+1}^h$ back to the central server.
\item After receiving the $M$ local updates $\theta_{t+1}^{m,h}$ from the clients, the server updates the global model $\theta_{t+1}$ by aggregating the results with the average function, layer-by-layer:
\begin{align*}
&\theta_{t+1} \longleftarrow \left\{\theta_{t+1}^{1}, \: \dots \: , \theta_{t+1}^{H}\right\}, \\
&\theta_{t+1}^{h} \longleftarrow \frac{1}{s} \sum_{m=1}^{s} \theta_{t+1}^{m,h}
\end{align*}
for $h=1, \dots, H$.
\end{itemize}

\section{Experimental Results}
\label{sec:results}
In this section, the proposed \alg{} algorithm with its \textit{full-network} and \textit{single-layer} implementations are evaluated from different points of view. The code of the following experiments, written in Python, can be found in the code repository\footnote{https://github.com/AI-Tech-Research-Lab/TIFeD}.

In the following experiments, we consider a fixed set of clients $M$, each with a local dataset $X_m$ uniformly sampled from the training set $X$. It is important to note that since we are in a tiny scenario where devices have a limited amount of memory (e.g., the Arduino Nano 33 BLE Sense board has 256kB of SRAM), the local dataset $X_m$ cannot be processed all at once as if it were entirely available to the device, because it could not fit into the memory of the device. Therefore, the local dataset $X_m$ is further divided into smaller portions of length $\textsf{buff\_len}$ and, at each round $t$, only one portion of the dataset of size $\textsf{buff\_len}$ is considered for training (i.e., the $t$-th buffer $X_m^t$).

\begin{table}[t]
\caption{NNs architecture considered in this experimental section. The rescaled Piecewise Linear Approximation (PLA) of the $\textit{tanh}$ function~\cite{song2022pocketnn} was used after each FC layer, whereas after the Conv2D layers the $\textit{ReLU}$ activation function was employed. The kernel size for the Conv2D layers is $\textbf{(3} \times \textbf{3)}$, while for the MaxPooling layer is $\textbf{(2} \times \textbf{2)}$. * Only the classifier is trained using the proposed \boldalg{} algorithm.}
\label{tab:models}
\begin{tabular}{ cc }
\toprule
    Model & Layers \\
    \toprule
    NN-1 & FC$(28*28) \rightarrow$ FC$(200) \rightarrow$ FC$(10)$ \\
    \midrule
    CNN-1 * & \begin{tabular}[l]{@{}l@{}}Conv2D$(4) \rightarrow$ MaxPool $\rightarrow$ Conv2D$(8) \rightarrow$ \\ MaxPool $\rightarrow$ FC$(200) \rightarrow$ FC$(50) \rightarrow$ FC$(10)$ \end{tabular} \\
    \midrule
    CNN-2 * & \begin{tabular}[l]{@{}l@{}}Conv2D$(128) \rightarrow$ MaxPool $\rightarrow$ Conv2D$(256) \rightarrow$ \\ MaxPool $\rightarrow$ Conv2D$(512) \rightarrow$ Conv2D$(256) \rightarrow$ \\ MaxPool $\rightarrow$ FC$(256) \rightarrow$ FC$(128) \rightarrow$ FC$(10)$ \end{tabular} \\
    \bottomrule
\end{tabular}
\end{table}

\subsection{Datasets}
\label{subsec:datasets}
A brief description of the datasets employed in our study is given in the following.

\subsubsection{MNIST}
\label{subsubsec:mnist}
The first considered dataset is the MNIST (Modified National Institute of Standards and Technology) database \cite{yann1998mnist}, a collection of handwritten digits commonly used for training and testing ML models able to recognize images. The elements are grey-scale images of size $28\times 28$. The dataset is composed of 60000 training images and 10000 testing images from 10 different classes, each one representing a digit from 0 to 9.

\subsubsection{FashionMNIST}
\label{subsubsec:f_mnist}
The FashionMNIST database \cite{xiao2017fashion} was created in 2017 as an improvement of the original MNIST dataset for benchmarking ML algorithms. FashionMNIST consists in gray-scale images of dimension $28\times 28$ of Zalando's articles. The training dataset contains 60000 elements, while the testing dataset includes 10000 elements. Each image is then associated with a label from 10 different classes.

\subsubsection{CIFAR10}
\label{subsubsec:cifar10}
Lastly, the CIFAR10 (Canadian Institute for Advanced Research) dataset \cite{krizhevsky2009cifar} is considered. CIFAR10 is a collection of colour images representing objects with size $32\times 32$ belonging to 10 different classes. The dataset is divided into 50000 training samples and 10000 test samples.

\begin{figure}[t]
     \centering
     \begin{subfigure}[b]{0.47\textwidth}
         \centering
         \includegraphics[width=\textwidth]{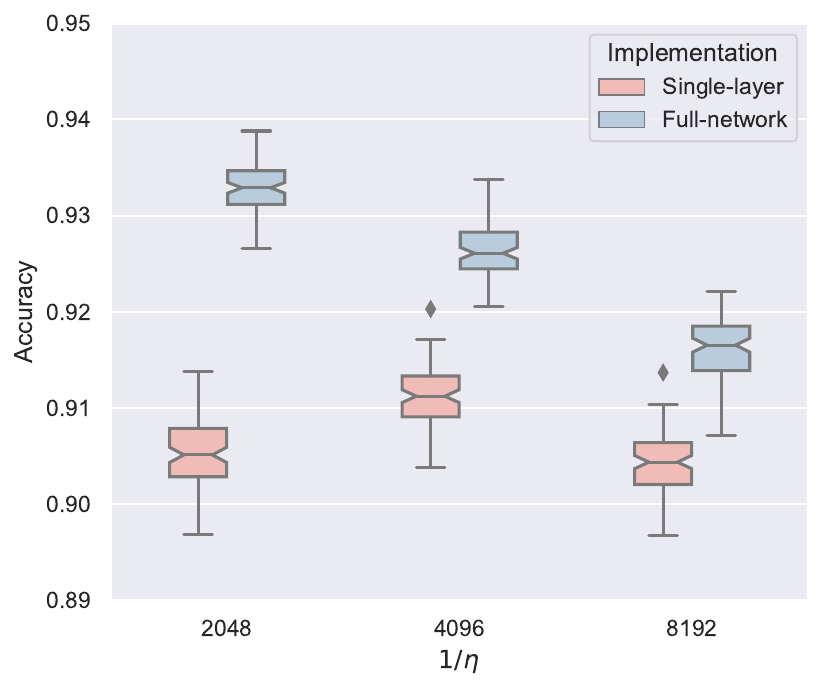}
         \caption{MNIST dataset}
     \end{subfigure}
     \hfill
     \begin{subfigure}[b]{0.47\textwidth}
         \centering
         \includegraphics[width=\textwidth]{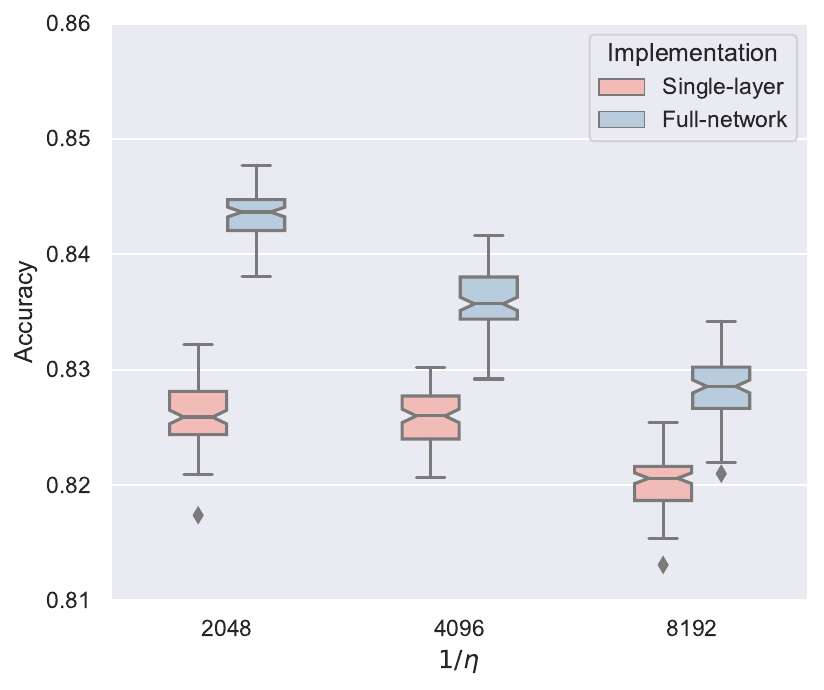}
         \caption{FashionMNIST dataset}
     \end{subfigure}
     \hfill
        \caption{Test accuracies of NN-1 trained with the proposed \textit{full-network} and \textit{single-layer} \alg{} implementations in a federated setting composed of $M=8$ worker nodes. NN-1 hyperparameters: length of the local dataset $X_m = 7.5k$ images, $\normalfont \textsf{buff\_len} = 50$, $bs = 25$, and $e = 5$. Three different values of the learning rate $\frac{1}{\eta}$ were considered. Results are computed for 100 different initialization of the DFA random-feedback weights $B^h$.}
        \label{fig:exp1}
\end{figure}

\subsection{Feed-forward neural network training}
\label{subsec:ffnn}
The goal of the first experiment is to show that the proposed \alg{} algorithm is able to train a feed-forward neural network from scratch. To this end, we trained a neural network composed of two fully-connected layers (NN-1 in Table~\ref{tab:models}) in a federated setting with $M=8$ worker nodes on the MNIST and FashionMNIST datasets with the following set of hyperparameters: length of the local dataset $X_m = 7.5k$ images, $\textsf{buff\_len} = 50$, $bs = 25$, and $e = 5$. We evaluated the test accuracy for both the \textit{full-network} and \textit{single-layer} implementations with three different values of the learning rate $\frac{1}{\eta}$ and for 100 different initialization of the DFA random-feedback weights $B^h$.

As we can see in Figure~\ref{fig:exp1}, the $M$ worker nodes are able to train a shared model initialized with zero-valued weights by using the proposed \alg{} algorithm. It is important to note that the \textit{single-layer} implementation performs worse than the \textit{full-network} implementation. This can be easily explained by the fact that, in the former case, each FC layer of the NN was trained by 4 out of 8 nodes, while, in the latter case, each NN layer is trained by all the 8 nodes in the federated setting. On the other hand, the computational, memory, and energy demands of the nodes are halved in the \textit{single-layer} implementation compared to the \textit{full-network} implementation, given that weight updates have to be computed and sent for only one layer $h$. As a consequence, the \textit{single-layer} implementation is more suitable when devices are extremely constrained in terms of resources.

\begin{figure}[t]
  \centering
        \begin{subfigure}[b]{0.47\textwidth}
        \centering
        \includegraphics[width=\textwidth]{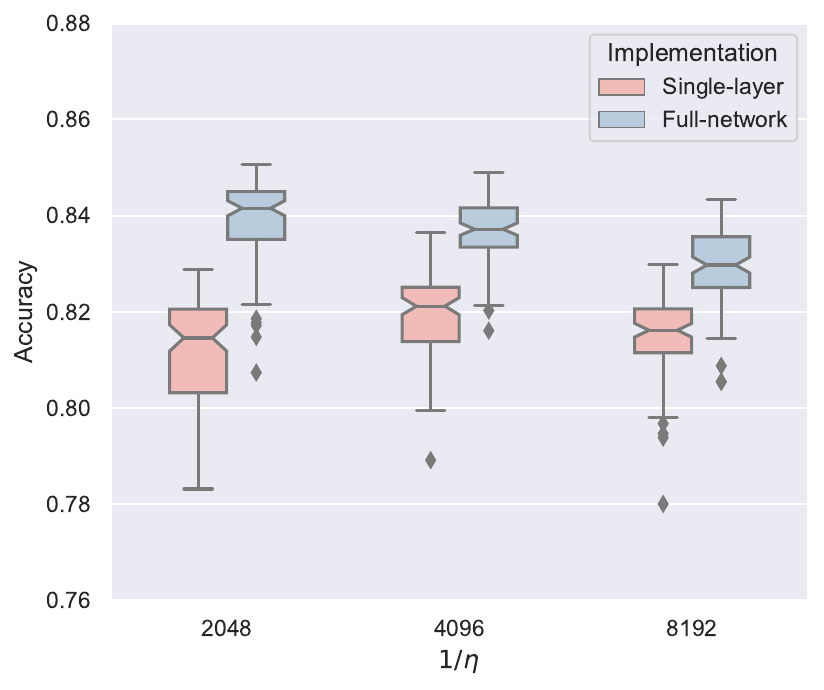}
    \end{subfigure}
    \caption{Test accuracies of CNN-1 trained, by exploiting transfer learning, with the proposed \textit{full-network} and \textit{single-layer} \boldalg{} implementations in a federated setting composed of $M=8$ worker nodes on the FashionMNIST dataset. CNN-1 hyperparameters: length of the local dataset $X_m = 7.5k$ images, $\normalfont \textsf{buff\_len} = 50$, $bs = 25$, and $e = 5$. Three different values of the learning rate $\frac{1}{\eta}$ were considered. Results are computed for 100 different initialization of the DFA random-feedback weights $B^h$.}
    \label{fig:exp2_fmnist}
\end{figure}

\subsection{Transfer learning for convolutional neural networks training}
\label{subsec:cnn}
In a TinyML environment, the ability to adapt a general and complex ML model to the specific application scenario is a fundamental aspect. This second experiment aims to prove that, in addition to feed-forward NNs, it is possible to use \alg{} to train the classifier of a Convolutional Neural Network (CNN) that exploits transfer learning.

First, we reconsidered the experiment presented in Section~\ref{subsec:ffnn} on the FashionMNIST dataset. We trained CNN-1 and quantized the feature extractor weights, which are then used by each worker node $m$. At this point, by exploiting TL, the FC layers of CNN-1 were re-trained from scratch directly on the worker nodes, using the same set of hyperparameters as in the experiment of Section~\ref{subsec:ffnn}. Similarly to the previous experiment, as can be seen from the results in Figure~\ref{fig:exp2_fmnist}, both the implementations of \alg{} are able to train the considered model.

\begin{figure}[t]
  \centering
        \begin{subfigure}[b]{0.47\textwidth}
        \centering
        \includegraphics[width=\textwidth]{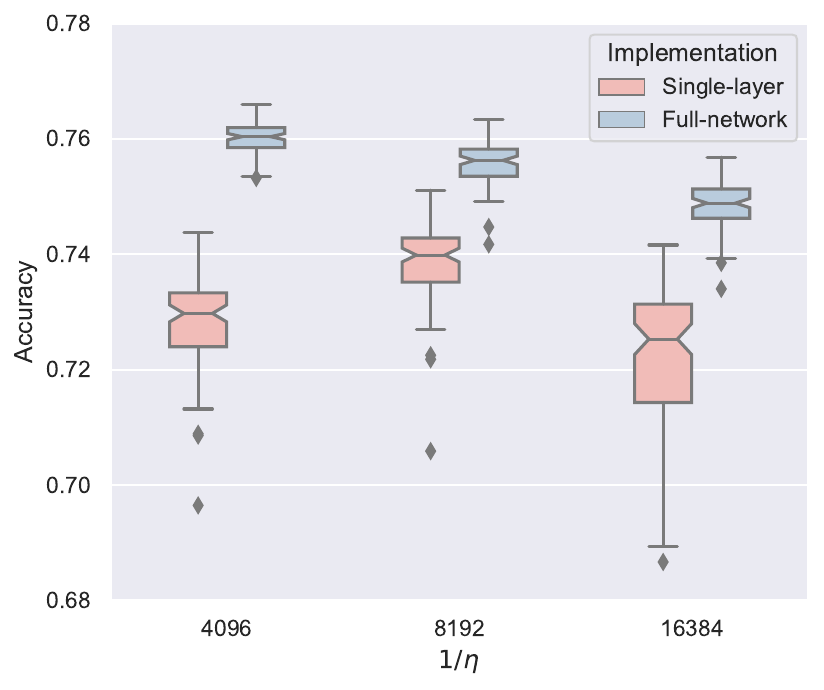}
    \end{subfigure}
    \caption{Test accuracies of CNN-2 trained, by exploiting transfer learning, with the proposed \textit{full-network} and \textit{single-layer} \boldalg{} implementations in a federated setting composed of $M=8$ worker nodes on the CIFAR10 dataset. CNN-2 hyperparameters: length of the local dataset $X_m = 6.25k$ images, $\normalfont \textsf{buff\_len} = 50$, $bs = 25$, and $e = 10$. Three different values of the learning rate $\frac{1}{\eta}$ were considered. Results are computed for 100 different initialization of the DFA random-feedback weights $B^h$.}
    \label{fig:exp2_cifar10}
\end{figure}

Then, by adopting the same procedure, we trained CNN-2 on the more complex CIFAR10 dataset. The following set of hyperparameters was used: length of the local dataset $X_m = 6.25k$ images, $\textsf{buff\_len} = 50$, $bs = 25$, and $e = 10$. The results, provided in Figure~\ref{fig:exp2_cifar10}, show that \alg{} is capable of making CNN-2 learn this more challenging image classification task, that a simpler feed-forward NN would not be able to solve.

\begin{figure}[t]
  \centering
        \begin{subfigure}[b]{0.47\textwidth}
        \centering
        \includegraphics[width=\textwidth]{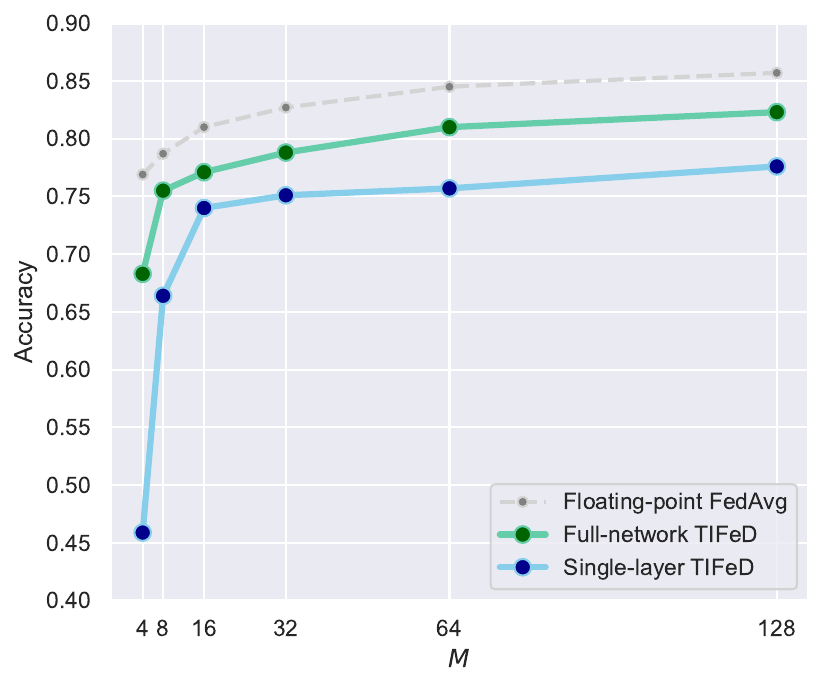}
    \end{subfigure}
    \caption{Analysis of the behaviour of \textit{full-network} \boldalg{} and \textit{single-layer} \boldalg{} as the number of worker nodes $M$ increases. Floating-point FedAvg is taken as a reference, but we emphasize that the goal is not to compare their final accuracies, since the two algorithms are tailored for two different application scenarios. Training of CNN-1, by exploiting transfer learning, on the FashionMNIST dataset. FedAvg hyperparameters: $X_m = 100$ images, $\normalfont \textsf{buff\_len} = 10$, $bs = 10$, $e = 10$, and $\eta = 0.1$. \boldalg{} hyperparameters: $X_m = 100$ images, $\normalfont \textsf{buff\_len} = 20$, $bs = 10$, $e = 10$, and $\frac{1}{\eta} = 2048$. Results are computed for 100 different initialization of the DFA random-feedback weights $B^h$ for \boldalg{} and 100 different network initialization for FedAvg.}
    \label{fig:exp3}
\end{figure}

\subsection{Exploring TIFeD as the number of worker nodes increases}
\label{subsec:fl}
In this third experiment, the behaviour of \alg{} as the number of worker nodes $M$ increases is analyzed. As a reference, we consider the state-of-the-art FL algorithm, i.e., Federated Averaging. We emphasize that this experiment is not intended to compare the final accuracy achieved by our \alg{} algorithm with respect to FedAvg. In fact, as previously mentioned, they are tailored for two deeply different application scenarios: \alg{} is specifically designed to run on resource-constrained devices, while FedAvg would not be able to be executed on tiny devices.

We trained the classifier of CNN-1 on the FashionMNIST dataset for different values of $M$. In particular, we set a fixed number of 100 images per worker and trained both FedAvg, which uses 32-bit long floating-point values, and the \textit{full-network} and \textit{single-layer} implementations of the \alg{} algorithm, which operate with 16-bit long integers and an integer-only arithmetic. The experiment was performed using the following set of hyperparameters for both algorithms: length of the local dataset $X_m = 100$ images, $bs = 10$, and $e = 10$. To take into account the lower memory consumption of \alg{} with respect to FedAvg, we set $\textsf{buff\_len} = 20$ for the former, $\textsf{buff\_len} = 10$ for the latter.

The results in Figure~\ref{fig:exp3} show that the \textit{full-network} and \textit{single-layer} \alg{} implementations behave exactly as FedAvg when increasing the number of worker nodes $M$. It should be noted that for the same number of nodes $M$, FedAvg performs twice as many communication rounds than the \textit{full-network} \alg{} implementation, since its buffer length is half as large, resulting in a longer and more energy-consuming training.

\subsection{The impact of the aggregation point within the TIFeD algorithm}
\label{subsec:fl}
Lastly, an experiment to evaluate the impact of the aggregation point within the \alg{} algorithm, is proposed. As explained in Section~\ref{subsec:feddfa}, the aggregation of the weights updates by the central server can occur at three different points: after each mini-batch $\alpha$, after the entire buffer $X_m^t$, or after the number of training epochs $e$. To this end, we performed the learning of the CNN-1 classifier with both the implementations of \alg{} on the FashionMNIST dataset, using the following set of hyperparameters: $M=128$, $X_m = 100$ images, $\textsf{buff\_len} = 20$, $bs = 10$, $e = 10$, and $\frac{1}{\eta} = 2048$.

As summarized in Table~\ref{tab:exp4}, the choice of the aggregation point represents a trade-off between the final test accuracy and the number of required communication rounds for both the \textit{full-network} and \textit{single-layer} implementations. In particular, the \textit{Epochs} aggregation point represents the one with the lowest number of required rounds, but it results in a final accuracy drop of 1.5\% and 1.8\% compared to that provided by the \textit{Buffer} aggregation point. On the other hand, the latter requires $e$ times more communication rounds than the former, resulting in more energy consumption by the tiny devices.

\begin{table}[t]
  \caption{Impact of the aggregation point within the \textit{full-network} and \textit{single-layer} \boldalg{} implementations in terms of test accuracy and number of communication rounds $t$. Training of CNN-1, by exploiting transfer learning, in a federated setting composed of $M=128$ worker nodes on the FashionMNIST dataset. CNN-1 hyperparameters: length of the local dataset $X_m = 100$ images, $\normalfont \textsf{buff\_len} = 20$, $bs = 10$, $e = 10$, and $\frac{1}{\eta} = 2048$. Results are computed for 100 different initialization of the DFA random-feedback weights $B^h$.}
  \label{tab:exp4}
\begin{tabular}{ cccc }
\toprule
    \begin{tabular}[c]{@{}c@{}}\alg{} \\ Implementation\end{tabular} & \begin{tabular}[c]{@{}c@{}}Aggregation \\ Point\end{tabular} & \begin{tabular}[c]{@{}c@{}}Test \\ Accuracy \end{tabular} & \begin{tabular}[c]{@{}c@{}} Communication \\ Rounds \end{tabular} \\ 
    \midrule
    \multirow{3}{*}{\textit{Full-network}} & Mini-batch $\alpha$ & 0.819 & 100\\
    & Buffer $X_m^t$ & 0.841 & 50\\
    & Epochs $e$ & 0.823 & 5\\
    \midrule
    \multirow{3}{*}{\textit{Single-layer}} & Mini-batch $\alpha$ & 0.780 & 100\\
    & Buffer $X_m^t$ & 0.791 & 50\\
    & Epochs $e$ & 0.776 & 5\\
    \bottomrule
\end{tabular}
\end{table}

\section{Conclusions}
\label{sec:conclusions}
The aim of this paper was to present a novel federated learning algorithm specifically designed for training ML and DL models directly on extremely resource-constrained devices. Specifically, we introduced \alg{}, a Tiny Integer-based Federated learning algorithm with Direct feedback alignment, with its two variants: the \textit{full-network} implementation, which allows each node in the federated setting to train all the fully-connected layers of the neural network, and the \textit{single-layer} implementation, that enables the devices to train only a single layer of the entire NN, resulting in a more lightweight algorithm in terms of computation, memory and energy required. The experimental results show the feasibility and effectiveness of the proposed solution. Future works will consider new models as well as further optimizations in the learning procedure.

%
\begin{acks}
This paper is supported by Dhiria s.r.l. and by PNRR-PE-AI FAIR project funded by the NextGeneration EU program.
\end{acks}
%
\bibliographystyle{ACM-Reference-Format}
\bibliography{main}


\begin{thebibliography}{19}


\ifx \showCODEN    \undefined \def \showCODEN     #1{\unskip}     \fi
\ifx \showDOI      \undefined \def \showDOI       #1{#1}\fi
\ifx \showISBNx    \undefined \def \showISBNx     #1{\unskip}     \fi
\ifx \showISBNxiii \undefined \def \showISBNxiii  #1{\unskip}     \fi
\ifx \showISSN     \undefined \def \showISSN      #1{\unskip}     \fi
\ifx \showLCCN     \undefined \def \showLCCN      #1{\unskip}     \fi
\ifx \shownote     \undefined \def \shownote      #1{#1}          \fi
\ifx \showarticletitle \undefined \def \showarticletitle #1{#1}   \fi
\ifx \showURL      \undefined \def \showURL       {\relax}        \fi
\providecommand\bibfield[2]{#2}
\providecommand\bibinfo[2]{#2}
\providecommand\natexlab[1]{#1}
\providecommand\showeprint[2][]{arXiv:#2}

\bibitem[Banbury et~al\mbox{.}(2020)]%
        {banbury2020benchmarking}
\bibfield{author}{\bibinfo{person}{Colby~R Banbury},
  \bibinfo{person}{Vijay~Janapa Reddi}, \bibinfo{person}{Max Lam},
  \bibinfo{person}{William Fu}, \bibinfo{person}{Amin Fazel},
  \bibinfo{person}{Jeremy Holleman}, \bibinfo{person}{Xinyuan Huang},
  \bibinfo{person}{Robert Hurtado}, \bibinfo{person}{David Kanter},
  \bibinfo{person}{Anton Lokhmotov}, {et~al\mbox{.}}}
  \bibinfo{year}{2020}\natexlab{}.
\newblock \showarticletitle{Benchmarking tinyml systems: Challenges and
  direction}.
\newblock \bibinfo{journal}{\emph{arXiv preprint arXiv:2003.04821}}
  (\bibinfo{year}{2020}).
\newblock


\bibitem[Bonawitz et~al\mbox{.}(2019)]%
        {bonawitz2019towards}
\bibfield{author}{\bibinfo{person}{Keith Bonawitz}, \bibinfo{person}{Hubert
  Eichner}, \bibinfo{person}{Wolfgang Grieskamp}, \bibinfo{person}{Dzmitry
  Huba}, \bibinfo{person}{Alex Ingerman}, \bibinfo{person}{Vladimir Ivanov},
  \bibinfo{person}{Chloe Kiddon}, \bibinfo{person}{Jakub Kone{\v{c}}n{\`y}},
  \bibinfo{person}{Stefano Mazzocchi}, \bibinfo{person}{Brendan McMahan},
  {et~al\mbox{.}}} \bibinfo{year}{2019}\natexlab{}.
\newblock \showarticletitle{Towards federated learning at scale: System
  design}.
\newblock \bibinfo{journal}{\emph{Proceedings of machine learning and systems}}
   \bibinfo{volume}{1} (\bibinfo{year}{2019}), \bibinfo{pages}{374--388}.
\newblock


\bibitem[Disabato and Roveri(2020)]%
        {disabato2020incremental}
\bibfield{author}{\bibinfo{person}{Simone Disabato} {and}
  \bibinfo{person}{Manuel Roveri}.} \bibinfo{year}{2020}\natexlab{}.
\newblock \showarticletitle{Incremental on-device tiny machine learning}. In
  \bibinfo{booktitle}{\emph{Proceedings of the 2nd International workshop on
  challenges in artificial intelligence and machine learning for internet of
  things}}. \bibinfo{pages}{7--13}.
\newblock


\bibitem[Goetz et~al\mbox{.}(2019)]%
        {goetz2019active}
\bibfield{author}{\bibinfo{person}{Jack Goetz}, \bibinfo{person}{Kshitiz
  Malik}, \bibinfo{person}{Duc Bui}, \bibinfo{person}{Seungwhan Moon},
  \bibinfo{person}{Honglei Liu}, {and} \bibinfo{person}{Anuj Kumar}.}
  \bibinfo{year}{2019}\natexlab{}.
\newblock \showarticletitle{Active federated learning}.
\newblock \bibinfo{journal}{\emph{arXiv preprint arXiv:1909.12641}}
  (\bibinfo{year}{2019}).
\newblock


\bibitem[Kairouz et~al\mbox{.}(2021)]%
        {kairouz2021advances}
\bibfield{author}{\bibinfo{person}{Peter Kairouz}, \bibinfo{person}{H~Brendan
  McMahan}, \bibinfo{person}{Brendan Avent}, \bibinfo{person}{Aur{\'e}lien
  Bellet}, \bibinfo{person}{Mehdi Bennis}, \bibinfo{person}{Arjun~Nitin
  Bhagoji}, \bibinfo{person}{Kallista Bonawitz}, \bibinfo{person}{Zachary
  Charles}, \bibinfo{person}{Graham Cormode}, \bibinfo{person}{Rachel
  Cummings}, {et~al\mbox{.}}} \bibinfo{year}{2021}\natexlab{}.
\newblock \showarticletitle{Advances and open problems in federated learning}.
\newblock \bibinfo{journal}{\emph{Foundations and Trends{\textregistered} in
  Machine Learning}} \bibinfo{volume}{14}, \bibinfo{number}{1--2}
  (\bibinfo{year}{2021}), \bibinfo{pages}{1--210}.
\newblock


\bibitem[Kone{\v{c}}n{\`y} et~al\mbox{.}(2016)]%
        {konevcny2016federated}
\bibfield{author}{\bibinfo{person}{Jakub Kone{\v{c}}n{\`y}},
  \bibinfo{person}{H~Brendan McMahan}, \bibinfo{person}{Felix~X Yu},
  \bibinfo{person}{Peter Richt{\'a}rik}, \bibinfo{person}{Ananda~Theertha
  Suresh}, {and} \bibinfo{person}{Dave Bacon}.}
  \bibinfo{year}{2016}\natexlab{}.
\newblock \showarticletitle{Federated learning: Strategies for improving
  communication efficiency}.
\newblock \bibinfo{journal}{\emph{arXiv preprint arXiv:1610.05492}}
  (\bibinfo{year}{2016}).
\newblock


\bibitem[Kopparapu and Lin(2021)]%
        {kopparapu2021tinyfedtl}
\bibfield{author}{\bibinfo{person}{Kavya Kopparapu} {and} \bibinfo{person}{Eric
  Lin}.} \bibinfo{year}{2021}\natexlab{}.
\newblock \showarticletitle{TinyFedTL: Federated transfer learning on tiny
  devices}.
\newblock \bibinfo{journal}{\emph{arXiv preprint arXiv:2110.01107}}
  (\bibinfo{year}{2021}).
\newblock


\bibitem[Krizhevsky et~al\mbox{.}(2009)]%
        {krizhevsky2009cifar}
\bibfield{author}{\bibinfo{person}{Alex Krizhevsky}, \bibinfo{person}{Vinod
  Nair}, {and} \bibinfo{person}{Geoffrey Hinton}.}
  \bibinfo{year}{2009}\natexlab{}.
\newblock \showarticletitle{Cifar-10 (canadian institute for advanced
  research). 2009}.
\newblock \bibinfo{journal}{\emph{URL http://www. cs. toronto. edu/kriz/cifar.
  html}}  \bibinfo{volume}{5} (\bibinfo{year}{2009}).
\newblock


\bibitem[Lillicrap et~al\mbox{.}(2014)]%
        {lillicrap2014random}
\bibfield{author}{\bibinfo{person}{Timothy~P Lillicrap},
  \bibinfo{person}{Daniel Cownden}, \bibinfo{person}{Douglas~B Tweed}, {and}
  \bibinfo{person}{Colin~J Akerman}.} \bibinfo{year}{2014}\natexlab{}.
\newblock \showarticletitle{Random feedback weights support learning in deep
  neural networks}.
\newblock \bibinfo{journal}{\emph{arXiv preprint arXiv:1411.0247}}
  (\bibinfo{year}{2014}).
\newblock


\bibitem[Llisterri~Gim{\'e}nez et~al\mbox{.}(2022)]%
        {llisterri2022device}
\bibfield{author}{\bibinfo{person}{Nil Llisterri~Gim{\'e}nez},
  \bibinfo{person}{Marc Monfort~Grau}, \bibinfo{person}{Roger Pueyo~Centelles},
  {and} \bibinfo{person}{Felix Freitag}.} \bibinfo{year}{2022}\natexlab{}.
\newblock \showarticletitle{On-device training of machine learning models on
  microcontrollers with federated learning}.
\newblock \bibinfo{journal}{\emph{Electronics}} \bibinfo{volume}{11},
  \bibinfo{number}{4} (\bibinfo{year}{2022}), \bibinfo{pages}{573}.
\newblock


\bibitem[McMahan et~al\mbox{.}(2017)]%
        {mcmahan2017communication}
\bibfield{author}{\bibinfo{person}{Brendan McMahan}, \bibinfo{person}{Eider
  Moore}, \bibinfo{person}{Daniel Ramage}, \bibinfo{person}{Seth Hampson},
  {and} \bibinfo{person}{Blaise~Aguera y Arcas}.}
  \bibinfo{year}{2017}\natexlab{}.
\newblock \showarticletitle{Communication-efficient learning of deep networks
  from decentralized data}. In \bibinfo{booktitle}{\emph{Artificial
  intelligence and statistics}}. PMLR, \bibinfo{pages}{1273--1282}.
\newblock


\bibitem[N{\o}kland(2016)]%
        {nokland2016direct}
\bibfield{author}{\bibinfo{person}{Arild N{\o}kland}.}
  \bibinfo{year}{2016}\natexlab{}.
\newblock \showarticletitle{Direct feedback alignment provides learning in deep
  neural networks}.
\newblock \bibinfo{journal}{\emph{Advances in neural information processing
  systems}}  \bibinfo{volume}{29} (\bibinfo{year}{2016}).
\newblock


\bibitem[Refinetti et~al\mbox{.}(2021)]%
        {refinetti2021align}
\bibfield{author}{\bibinfo{person}{Maria Refinetti},
  \bibinfo{person}{St{\'e}phane d’Ascoli}, \bibinfo{person}{Ruben Ohana},
  {and} \bibinfo{person}{Sebastian Goldt}.} \bibinfo{year}{2021}\natexlab{}.
\newblock \showarticletitle{Align, then memorise: the dynamics of learning with
  feedback alignment}. In \bibinfo{booktitle}{\emph{International Conference on
  Machine Learning}}. PMLR, \bibinfo{pages}{8925--8935}.
\newblock


\bibitem[Song and Lin(2022)]%
        {song2022pocketnn}
\bibfield{author}{\bibinfo{person}{Jaewoo Song} {and} \bibinfo{person}{Fangzhen
  Lin}.} \bibinfo{year}{2022}\natexlab{}.
\newblock \showarticletitle{PocketNN: Integer-only Training and Inference of
  Neural Networks via Direct Feedback Alignment and Pocket Activations in Pure
  C++}.
\newblock \bibinfo{journal}{\emph{arXiv preprint arXiv:2201.02863}}
  (\bibinfo{year}{2022}).
\newblock


\bibitem[Stankovic(1996)]%
        {stankovic1996real}
\bibfield{author}{\bibinfo{person}{John~A Stankovic}.}
  \bibinfo{year}{1996}\natexlab{}.
\newblock \showarticletitle{Real-time and embedded systems}.
\newblock \bibinfo{journal}{\emph{ACM Computing Surveys (CSUR)}}
  \bibinfo{volume}{28}, \bibinfo{number}{1} (\bibinfo{year}{1996}),
  \bibinfo{pages}{205--208}.
\newblock


\bibitem[Sundaramoorthy et~al\mbox{.}(2018)]%
        {sundaramoorthy2018harnet}
\bibfield{author}{\bibinfo{person}{Prahalathan Sundaramoorthy},
  \bibinfo{person}{Gautham~Krishna Gudur}, \bibinfo{person}{Manav~Rajiv
  Moorthy}, \bibinfo{person}{R~Nidhi Bhandari}, {and} \bibinfo{person}{Vineeth
  Vijayaraghavan}.} \bibinfo{year}{2018}\natexlab{}.
\newblock \showarticletitle{Harnet: Towards on-device incremental learning
  using deep ensembles on constrained devices}. In
  \bibinfo{booktitle}{\emph{Proceedings of the 2nd International Workshop on
  Embedded and Mobile Deep Learning}}. \bibinfo{pages}{31--36}.
\newblock


\bibitem[Xiao et~al\mbox{.}(2017)]%
        {xiao2017fashion}
\bibfield{author}{\bibinfo{person}{Han Xiao}, \bibinfo{person}{Kashif Rasul},
  {and} \bibinfo{person}{Roland Vollgraf}.} \bibinfo{year}{2017}\natexlab{}.
\newblock \showarticletitle{Fashion-mnist: a novel image dataset for
  benchmarking machine learning algorithms}.
\newblock \bibinfo{journal}{\emph{arXiv preprint arXiv:1708.07747}}
  (\bibinfo{year}{2017}).
\newblock


\bibitem[Yang et~al\mbox{.}(2020)]%
        {yang2020federated}
\bibfield{author}{\bibinfo{person}{Kai Yang}, \bibinfo{person}{Tao Jiang},
  \bibinfo{person}{Yuanming Shi}, {and} \bibinfo{person}{Zhi Ding}.}
  \bibinfo{year}{2020}\natexlab{}.
\newblock \showarticletitle{Federated learning via over-the-air computation}.
\newblock \bibinfo{journal}{\emph{IEEE Transactions on Wireless
  Communications}} \bibinfo{volume}{19}, \bibinfo{number}{3}
  (\bibinfo{year}{2020}), \bibinfo{pages}{2022--2035}.
\newblock


\bibitem[Yann(1998)]%
        {yann1998mnist}
\bibfield{author}{\bibinfo{person}{LeCun Yann}.}
  \bibinfo{year}{1998}\natexlab{}.
\newblock \showarticletitle{The mnist database of handwritten digits}.
\newblock \bibinfo{journal}{\emph{R}} (\bibinfo{year}{1998}).
\newblock


\end{thebibliography}

\end{document}